\documentclass[letterpaper, 10 pt, conference， twocolumn]{ieeeconf}  

\IEEEoverridecommandlockouts                              

\usepackage{mathptmx} 
\usepackage{amsmath} 
\usepackage{amssymb}  

\usepackage{algorithm} 
\usepackage{subcaption}
\usepackage{graphicx}
\usepackage{algpseudocode}
\usepackage{float} 
\usepackage[unicode=true,pdfusetitle,
 bookmarks=true,bookmarksnumbered=false,bookmarksopen=false,
 breaklinks=false,pdfborder={0 0 1},backref=false,colorlinks=true,
 linkcolor=blue, citecolor=blue, urlcolor=black]
 {hyperref}
\usepackage{cite}

 \usepackage{babel}
\usepackage[latin9]{inputenc}
\usepackage{caption}
\usepackage{makecell}
\usepackage{booktabs}
\usepackage{multirow}
\usepackage{xcolor} 

\makeatother

\newif \iffinal
\finalfalse
\iffinal
\newcommand{\WJmodified}[1]{{#1}}
\newcommand{\WJcomments}[1]{{}}
\newcommand{\QXYModified}[1]{{#1}}
\newcommand{\QXYComments}[1]{{}}
\newcommand{\oldText}[1]{}
\newcommand{\toAppendix}[1]{}
\else
\newcommand{\WJmodified}[1]{{\color{black} #1}}
\newcommand{\WJcomments}[1]{{\color{black} \WJmodified{Wenjie commented: #1}}}
\newcommand{\QXYModified}[1]{{\color{orange} #1}} 
\newcommand{\QXYComments}[1]{{\color{purple} \QXYModified{QXY commented: #1}}}
\newcommand{\toAppendix}[1]{#1}
\newcommand{\oldText}[1]{#1}
\fi

\title{\LARGE \bf
\WJmodified{Data-driven inventory management for new products: An adjusted Dyna-$Q$ approach with transfer learning}
}

\author{Xinye Qu$^{1}$, Longxiao Liu$^{2}$ and Wenjie Huang$^{3}$
\thanks{*This work was supported by National Science Foundation of China under Grant 72201224 and Hong Kong Research Grants Council Theme-based Research Scheme under Grant T32-707-22-N.}
\thanks{$^{1,2}$ Xinye Qu ({\tt\small xinyequ@connect.hku.hk}) and Longxiao Liu ({\tt\small liulx@connect.hku.hk}) are with Department of Data and Systems Engineering, The University of Hong Kong, Pokfulam Road, Hong Kong SAR.
       }%
\thanks{$^{3}$ Wenjie Huang ({\tt\small huangwj@hku.hk}) is with Department of Data and Systems Engineering and Musketeers Foundation Institute of Data Science, The University of Hong Kong, Pokfulam Road, Hong Kong SAR.
       }%
}

\begin{document}

\maketitle
\thispagestyle{empty}
\pagestyle{empty}

\begin{abstract}
In this paper, we propose a novel reinforcement learning algorithm for inventory management of newly launched products with no historical demand information. The algorithm follows the classic Dyna-$Q$ structure, balancing the model-free and model-based approaches, while accelerating the training process of Dyna-$Q$ and mitigating the model discrepancy generated by the model-based feedback. Based on the idea of transfer learning, warm-start information from the demand data of existing similar products can be incorporated into the algorithm to further stabilize the early-stage training and reduce the variance of the estimated optimal policy. Our approach is validated through a case study of bakery inventory management with real data. The adjusted Dyna-$Q$ shows up to a 23.7\% reduction in average daily cost compared with $Q$-learning, and up to a 77.5\% reduction in training time within the same horizon compared with classic Dyna-$Q$. By using transfer learning, it can be found that the adjusted Dyna-$Q$ has the lowest total cost, lowest variance in total cost, and relatively low shortage percentages among all the benchmarking algorithms under a 30-day testing.
\end{abstract}

\section{INTRODUCTION}

Inventory management is crucial for supply chain operations, overseeing and controlling the order, storage, and usage of goods in business \cite{haijema2014optimal}. In inventory management, the cold-start setting refers to predicting demand and formulating appropriate inventory strategies when new products are introduced or new market demands arise due to the lack of historical data \cite{hung2022addressing}. Data-driven flexible inventory management strategies are adopted, such as Just-in-Time inventory management \cite{simchi2018increasing}, inventory optimization models \cite{poirier1996supply}, and predictive analytics and intelligent methods \cite{giannoccaro2002inventory}.

Reinforcement learning (RL) has been increasingly applied in inventory management to optimize inventory control policies and decision-making processes \cite{giannoccaro2002inventory}. Most research on RL's application in inventory management mainly addresses the problems with existing similar products, whose historical sales information is accessible \cite{pontrandolfo2002global}. The RL method's agent has adequate data to obtain a satisfactory ordering strategy \cite{demizu2023inventory}. Most studies apply model-free RL in inventory control under sufficient data due to model-free RL's satisfactory performance under stochastic conditions \cite{boute2022deep}. On the other hand, model-based RL has the potential to improve upon model-free methods by offering more stable and interpretable policies and potentially faster convergence. However, for newly launched products without actual demand data, the demand prediction model is trained only based on the historical demands of existing products \cite{demizu2023inventory}. The discrepancy between the trained model and the actual environment may prevent the model-based algorithm from converging toward the near-optimal policy \cite{wang2020offline}. Consequently, incorporating demand information from the actual environment obtained by model-free RL methods could benefit the model-based algorithm by obtaining a better result. Therefore, both model-free RL and model-based RL have demonstrated significant potential in optimizing inventory management policies and decision-making processes. However, challenges persist in how to merge their merits together in inventory management for new products with no historical demand data.

The Dyna-$Q$ algorithm combines model-based RL with model-free RL, which learns from a continuously updated model and the real environment, to update the value function or policy function \cite{sutton1991dyna}. This approach helps address inventory forecasting and management challenges when historical data is unavailable. By utilizing simulation and planning, Dyna-$Q$ compensates for the absence of actual data, allowing it to learn and adapt to new environments incrementally.
Simulating an environment model that accurately reflects the dynamics of the real environment is challenging, especially in the early stages of training. If the model deviates from the actual environment, policy updates based on the simulated data may lead to suboptimal results. Even in cases where a relatively accurate model can be simulated, maintaining this model demands significant computational resources and data, particularly in complex or dynamically changing environments \cite{pei2021improved, zou2020pseudo}.

\WJmodified{There exist technologies to enhance learning efficiency and performance through the application of knowledge acquired from one task to another related task, which constitutes a form of transfer learning. Transfer $Q$-learning frequently expedites the learning process for a new task by transferring knowledge of models, $Q$-tables, targets, features, and parameters from a source task \cite{weiss2016survey}. An Imitation and Transfer $Q$-learning (ITQ) approach has been proposed for load parameter identification problem \cite{xie2020imitation}, which transfers the $Q$-value from the source task as the initial start of the new task. 
ITQ effectively balances greedy and global search strategies, thereby preventing premature convergence and attaining global optimum, with simulation results demonstrating superior convergence compared to traditional methodologies. Target transfer $Q$-learning has been introduced to transfer the $Q$-function from a source task as the temporal difference target for a new task under safe conditions, ensuring more rapid convergence than standard $Q$-learning \cite{wang2020target}. Comprehensive task models were transferred in hybrid RL agents utilizing inter-task mappings and cascade neural networks, amalgamating model-free and model-based learning to achieve substantial performance enhancements over non-transfer approaches \cite{fachantidis2011transferring}. 
Empirical results confirm its 43.58\% improvement in efficiency relative to conventional deep transfer learning methods, representing the inaugural study to optimize knowledge transfer for deep learning in networking applications \cite{phan2020q}. To the best of our knowledge, no research has been done that integrates transfer learning with Dyna-$Q$ within the context of inventory control problems. There is also little work on transferring both $Q$-value and the model of a source task in model-based RL.}

 
\WJmodified{To address the existing limitations mentioned above, this study firstly introduces an adjusted Dyna-$Q$ algorithm, which incorporates the search-then-convergence (STC) process to design the iteration-decaying exploration and planning steps, dynamically reweighting the model-based and model-free components. Secondly, based on the idea of transfer learning, a demand forecasting model is constructed based on the demand for existing similar products to provide warm-start prior knowledge of the new product's demand distribution for the adjusted Dyna-$Q$ algorithm. Finally, the proposed approach is validated using real data from a bakery to manage inventory for a newly launched product. The results demonstrate the superior performance of the proposed approach in terms of cost and variance reduction compared to classic Dyna-$Q$ and $Q$-learning algorithms.}
\section{Framework of the Proposed Approach}
This section introduces the proposed algorithm, which is \WJmodified{developed} based on the classic Dyna-$Q$ method and aims to minimize long-term discounted operation costs by determining the optimal ordering quantity when the demand distribution of a new product is unknown. The proposed algorithm offers two main innovations. Firstly, the selection probability of arbitrary actions in the epsilon-greedy strategy and the number of planning steps are determined using the search-then-convergence (STC) process \cite{darken1992learning}, which decays with the iteration over the training horizons. This approach accelerates the training process and places greater emphasis on exploiting the model-free components as more experience is gathered from the environment. Secondly, since there is no historical demand data for the new product during the early training phases, the proposed algorithm incorporates transfer learning by utilizing simulated data from existing similar products as warm-start information. These two innovations will be further elaborated in the subsequent subsections.

 Denote state (the inventory levels) space $\mathbb{S}$ and for any $s\in \mathbb{S}$, an action (the ordering quantities) space $\mathbb{A}(s)$. Assume that the state and action spaces are all finite. The proposed Algorithm \ref{adjusted Dyna-$Q$ algorithm} follows the structure of Dyna-$Q$, which combines both model-free and model-based methods. In each iteration $t$ within the training horizons, the algorithm directly interacts with the environment and updates the $Q$-values following model-free $Q$-learning. For a given state-action pair $s\in\mathbb{S},\,a\in\mathbb{A}(s)$, 

\begin{equation}
    \label{update_q_table}
    Q(s,\,a)\leftarrow Q(s,\,a)+\alpha[c+\gamma \min_{a^\prime\in\mathbb{A}(s)}Q(s^\prime,\,a^\prime)-Q(s,\,a)],
\end{equation}
where $c$ denotes the immediate cost, $s^\prime \in \mathbb{S}$ the next state transited, $0<\alpha<1$ the learning rate, and $0<\gamma<1$ the discount factor. The new experience $c$ and $s^\prime$ are used in updating the model. Here, we use $M(s,\,a)$ to denote the model associated with state-action pair $(s,\,a)$. 
Explicitly, the model $M(s,\,a)$ consists of two neural networks estimating the state transition probabilities $P(s'|s,\,a)$ and the cost function $C(s,\,a,\,s')$, respectively. The estimation is updated by adjusting the parameters to minimize the least squares error between the predicted state transition and cost and the new experience \cite{polydoros2017survey}.
In terms of the model-based part, multiple planning steps are conducted based on the model simulation, where the $Q$-table is further updated, using the same rule as formulation (\ref{update_q_table}).

\subsection{Adjusted Exploration Strategy and Planning Steps based on Search-then-Convergence (STC) Process}
An epsilon-greedy strategy has been implemented in each iteration to balance the exploitation and exploration processes. In Algorithm \ref{adjusted Dyna-$Q$ algorithm}, we adjust $\epsilon_t$ based on the search-then-convergence (STC) process \cite{darken1992learning}.
In the early iterations of training, the agent tends to explore more rather than exploit the current best action. As iteration evolves, the epsilon value will decrease to a specific value, and the agent will emphasize more on exploitation. The epsilon value is updated based on the STC procedure as below:
$\varepsilon_t=\max\left\{\varepsilon_0/1+y,\,\varepsilon_{\min}\right\}$,
where
    $y=t^2/(\partial_\varepsilon+t)$.
Here $\varepsilon_0>0$ represents the initial value of the epsilon, $\varepsilon_{\min}>0$ represents the lower bound of the epsilon, and $\partial_\varepsilon$ is a constant smoothing the decaying process, which will be chosen appropriately in practice.

The planning component (model-based part) can help the agent examine and update the $Q$-table by exploiting past experiences from the environment. It is known for reducing the sample complexity, leading to faster convergence and variance reduction \cite{lowrey2018plan}\cite{luo2018algorithmic}. 
However, a trade-off exists as it is less computationally efficient than model-free $Q$-learning within the same training horizons.
Algorithm \ref{adjusted Dyna-$Q$ algorithm} enables acceleration by setting the total planning steps decaying with the iteration, which means that the model-free part is more emphasized as the iteration evolves. The intention is that in the late training horizons, the agent has obtained enough exploration for the environment and can rely on model-free part, which is computationally efficient. The planning steps $N_t$ are updated based on the STC procedure as well:
    $N_t=\max\left\{N_0/1+y,\,N_{\min} \right\},$
where
    $y=t^2/(\partial_N+t)$.
Here, $N_0>0$ represents the initial value of the planning steps, $N_{\min}>0$ represents the lower bound of the planning steps, and $\partial_N$ is a constant smoothing the decaying process, which will be chosen appropriately in practice. 

The anticipated benefit of the adjusted Dyna-$Q$ algorithm is that it accelerates the training process while achieving lower costs in both the training and testing phases. 

\begin{algorithm*} [t]
\caption{Dyna-$Q$ with Adjusted Exploration Strategy and Planning Steps}
\label{adjusted Dyna-$Q$ algorithm}
\begin{algorithmic}
\Statex Initialize $Q(s, a)$ arbitrarily, Initialize the model $M(s, a)$ for environment simulation, for all state-action pairs. Set the size of the training horizon $T$, the learning rate $\alpha$ and the discount factor $\gamma$.
\Statex \textbf{When iteration $t \leq T$:}
    \Statex \qquad Initialize state $s$
    \Statex \qquad Update \textbf{epsilon} $\epsilon_t$ and \textbf{planning steps} $N_t$ based on STC process.
        \Statex \qquad Choose action $a$ from state $s$ using policy derived from $Q$ (e.g., $\epsilon_t$-greedy)
        \Statex \qquad Take action $a$, observe cost $c$ and next state $s^\prime$  
         \Statex \qquad Update $Q(s,\,a)$ using the rule (\ref{update_q_table})
        \Statex \qquad $M(s,\,a) \gets (s^\prime,\,c)$ \Comment{\emph{Update the model with the new experience}}
        \State \qquad $s \gets s^\prime$  \Comment{\emph{Move to the new state}}
            \Statex \qquad\textbf{When planning step $n\leq N_t$:}  \Comment{\emph{Planning: Learning from simulated experience}}
            \Statex \qquad \qquad Randomly pick a previously observed state $s$ and an action taken under such state $a$
            \Statex \qquad\qquad Retrieve simulated next state $s^\prime$ and cost $c$ from $M(s,\, a)$
            \Statex \qquad\qquad Update $Q(s,\,a)$ using the rule (\ref{update_q_table})
\end{algorithmic}
\end{algorithm*}
\subsection{Warm-start Information from Transfer Learning}
Algorithm \ref{adjusted Dyna-$Q$ algorithm} operates as a cold-start algorithm without prior demand information for new products. This absence of demand information at the beginning can cause discrepancies between the estimated model and the actual environment in the early training phases, potentially resulting in suboptimal policy estimates. To address this issue, another advanced variant of the algorithm with transfer learning is developed, integrating a simulation-based demand forecasting model to provide warm-start knowledge of new product demand for the adjusted Dyna-$Q$ algorithm. The demand forecasting model is structured as a three-layer Bayesian neural network (BNN), trained on historical data of similar existing products. BNN can leverage prior knowledge, such as product sales seasonality and market trends, to offer improved predictions of new product demand \cite{jospin2022hands}. This demand forecasting model, denoted as $f_{\theta}(u,v)$, utilizes numerical values related to historical demands and moving averages within a specific window size denoted by $u$, along with multiple categorical features related to calendar data for the day before the demand, marked as $v$. These features include: day of the week, weekday/weekend indicator, week number, and sine and cosine values representing the days elapsed in the month and year. Here, $\theta$ represents the model parameters. By leveraging the demand forecasting model, an offline simulated demand dataset $D = [\hat{d}_1,...,\hat{d}_h]$ can be generated by inputting the corresponding day features, with $h$ representing the prediction horizon.

\begin{figure*}[htbp]
\centering
\includegraphics[scale=0.45]{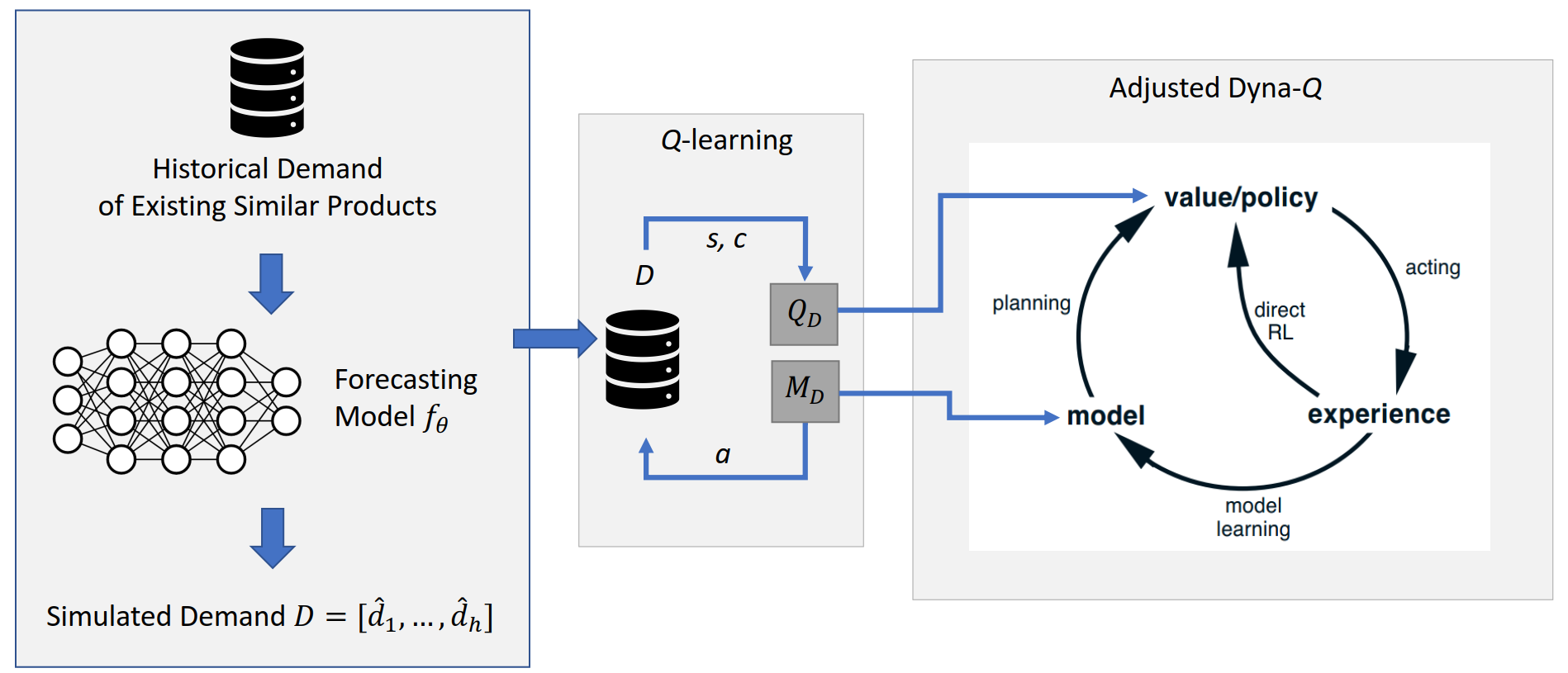}
\caption{Pipeline of adjusted Dyna-$Q$ with transfer learning}
\label{fig4.1}
\end{figure*}

The transfer learning process is executed as follows. Before implementing Algorithm \ref{adjusted Dyna-$Q$ algorithm}, we deploy $Q$-learning on the offline data set $D$ and obtain an updated $Q$-table, denoted as $Q_D$, which will be transferred as the initial $Q$-table in Algorithm \ref{adjusted Dyna-$Q$ algorithm}. 
The updated model $M_D$ is transferred as the initialization for Algorithm \ref{adjusted Dyna-$Q$ algorithm}. Figure \ref{fig4.1} provides a concise overview of the structure of the adjusted Dyna-$Q$ approach with transfer learning. Incorporating transfer learning into Algorithm \ref{adjusted Dyna-$Q$ algorithm} allows the simulated demand data from similar existing products to stabilize the demand distribution shift of the new product in the early training phases, thereby reducing training variance. As more new experiences are gathered, the discrepancy between the demand distributions of existing and new products can be mitigated. The adjusted approach introduced in Section II.$A$ can further aid in minimizing this discrepancy.
\section{Case Study}
In this case study, we apply the proposed algorithms to address a real-life single-echelon inventory problem faced by a bakery. The bakery store conducts periodic inventory reviews and places orders from a single, reliable supplier with an unlimited supply capacity. The inventory review and reorder period are set to one day, assuming that the bakery receives the ordered products immediately. Given the perishable nature of the products, it is assumed that the quality decay process begins upon immediate receipt by the store and lasts for three days. Any unsold products held in inventory for three days will be discarded. The bakery has introduced a new product called Boule 400g (Boule, from French, meaning ''ball'', is a traditional shape of French bread resembling a squashed ball), with its daily demand assumed to follow a random and identically distributed pattern. A dataset containing daily transaction details of customers from a bakery in France, collected from Kaggle \cite{Gimbert_2022}, provides the daily transaction information from 2021-01-01 to 2022-09-3 for the existing similar product: Boule 200g. The historical demand data for Boule 200g from the bakery can be leveraged to generate warm-start information for the new product, Boule 400g.

The sequence of the bakery's behaviors is elaborated as follows. At the beginning of each day, the store receives the orders placed yesterday, and the inventory level is updated. Then, the products are sold on a FIFO (First-in and first-out) rule and unfulfilled demand is considered a loss of sales. Next, the quality of products is monitored, and the products lasting in inventory for three days will be discarded. Finally, the inventory level is updated again, and the store places replenishment orders. We formulate the problem and environment in the following subsection.

\subsection{Problem Formulation and Environment Construction}
The state variable contains vital information on inventory levels. The state at period $t$ is shown as:  $s_t=\{s^1_t,\,s^2_t,\,s^3_t\}$, where $s^i_t$ represents the inventory levels of products that have $i$ days shelf-life. For instance, $s^3_t$ denotes the inventory level of products with a 3-day shelf-life, the fresh products the store just received that day. The state $s^2_t$ and $s^1_t$ denote the inventory level of products with 2 days or 1-day shelf-life, respectively. At the end of period $t$, the products with $s^1_t$ will be disposed of from the inventory. Note that $s^1_t,s^2_t$ and $s^3_t$ are all integers.

The bakery's action $a_t$ denotes the order quantity of products at the end of period $t$. Suppose that the maximum possible inventory level $|\mathbb{S}|$ is predetermined and the quantity $a_t$ is an integer with a proper upper bound depending on the maximum possible demand and inventory level. In each period, the total cost $C_t$ is represented as
\begin{align}
     C_t= & b_3\times s^3_t+b_2\times s^2_t+b_1\times s^1_t \nonumber
     \\
     & +C_s \max\{d_t-s^1_t-s^2_t-s^3_t,\,0\}, \label{cost}
\end{align}
where $d_t$ denotes the observed demand at period $t$, $C_s$ represents the unit cost for the shortage of unfulfilled demand, and $b_i$ represents the holding/obsoleting inventory costs for products with shelf-life of $i$ day(s). We assume that $b_1 > b_2 \geq b_3$ by problem natural since these deteriorated products may result in additional loss of sales due to sales discounts. And the disposition of products $s^1_t$ will be charged extra costs. Without loss of generality, we assume that there is no ordering cost. 

For our problem, two state transition processes will occur: one at the beginning of the period and the other at the end associated with customer demand. The first state transition is elaborated in the formula as $s^1_t=s^2_{t-1}$, $s^2_t=s^3_{t-1}$, $s^3_t=a_{t-1}$, which summarizes that the shelf-life of all products will decrease by one due to decaying. Then the $s_t$ will be used to compute the cost by equation (\ref{cost}). The second state transition occurs after the bakery interacts with the environment, returning daily demand information $d_t$:
\begin{equation}
      \label{3.9}
      s^3_t=\left\{
      \begin{aligned}
         &s^3_t, & \text{if }d_t<s^1_t+s^2_t,\\
         &\max\{s^3_t-(d_t-s^1_t-s^2_t),\,0\}, & \text{otherwise}. 
      \end{aligned}
     \right.
\end{equation}
\begin{equation}
      \label{3.10}
      s^2_t=\left\{
      \begin{aligned}
         &s^2_t,& \text{if } d_t<s^1_t,\\
         &\max\{s^s_t-(d_t-s^1_t),\,0\}, & \text{otherwise}. 
      \end{aligned}
     \right.
\end{equation}
\begin{equation}
    \label{3.11}
    s^1_t=\max\{s^1_t-d_t,\,0\}.
\end{equation}
The state transition processes (\ref{3.9})-(\ref{3.11}) will be updated simultaneously and follows the FIFO rule, that the demand will be first served with inventory with the least shelf-life, which are the products with $s^1_t$. If the demand can not be fulfilled by $s^1_t$, then $s^2_t$ and then $s^3_t$ will be consumed. 

The goal of the inventory management problem is to find an optimal stationary policy (optimal $a_t$ given state $s_t$ at each period), such that the long-term expected cost $\mathbb{E}[\sum_{t=1}^{\infty} \gamma^{t-1} C_t]$ is minimized.

\subsection{Implementation and Result Discussions}
Set the unit cost parameters as: $[b1,\,b2,\,b3] = [0.7,\,0.3,\,0]$, $C_s = 1$, all in Euro. \WJmodified{By analyzing the demand distribution of the existing product, Boule 200g, we find the mean and maximum daily demand is 4.48 and 10, respectively. 
So we assume that the daily demand of the new product: Boule 400g follows a discretize distribution on $[0,\,10]$ close to a Gamma distribution with mean $\mu = 5$ and with three different variance settings $\sigma^2 = 1,3$ or $5$.} The different variance settings can be achieved by adjusting the shape and scale parameters of the Gamma distribution. The initial inventory level is $s_1 = (0,\,0,\,5)$. All experiments were run on a high-performance server with NVIDIA Tesla T4 GPUs.

We first apply the adjusted Dyna-$Q$ algorithm (without transfer learning), with parameters $\alpha=0.3,\,\gamma=0.9,\epsilon_0 = 0.4,\,\epsilon_{\min} = 0.1,\,\,N_0 = 100,\,N_{\min} = 10,\,\partial_{\epsilon}=7500,\,\partial_{N} = 5000$. For comparison, we construct the model $M(s,\,a)$ by two either BNNs or Multi-Layer Perceptrons (MLPs) (one for cost function and another for state transition probability). 
Each BNN comprises an input layer, two hidden layers (with 128 and 64 neurons, both utilizing ReLU activation), and a linear output layer, with dropout layers (with dropout probability $0.5$) applied after each hidden layer to approximate Bayesian inference. The model is optimized using the Adam optimizer (with learning rate 0.001) with mean squared error (MSE) as the loss function, and during testing, prediction uncertainty is quantified through 10 Monte Carlo sampling iterations to compute both the mean prediction and variance. 
The MLP maintains an identical architecture 
as BNN, but omits the Dropout layers to establish a deterministic model, while similarly employing the Adam optimizer and MSE loss function. The MLP's parameters are optimized via standard backpropagation, with predictions generated through a single forward pass \cite{popescu2009multilayer}.

\begin{figure}[htbp]
\centering
\includegraphics[scale=0.37]{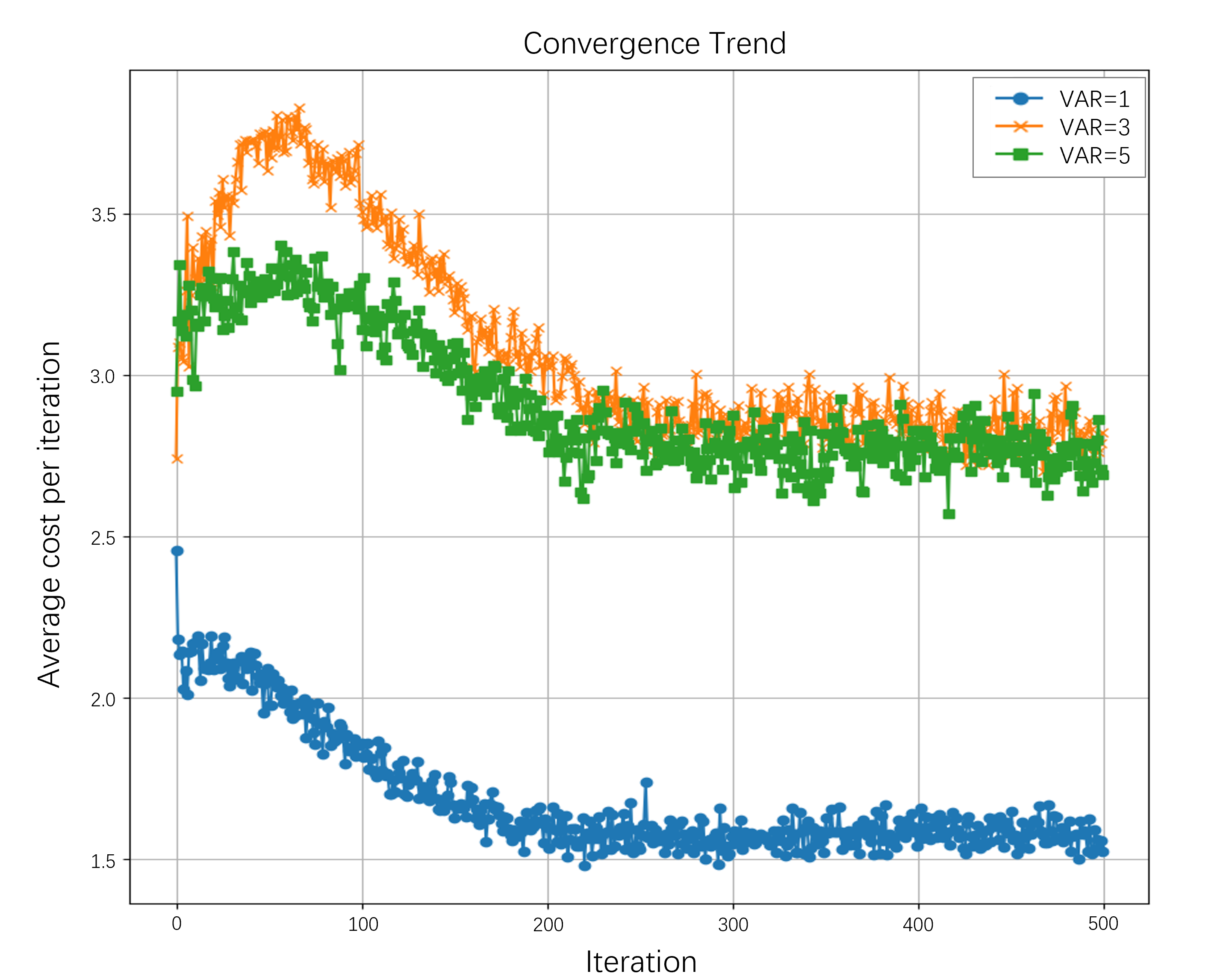}
\caption{Convergence of adjusted Dyna-$Q$ algorithm (``Var'' represents the variance $\sigma^2$)}
\label{figure2new1}
\end{figure}

Our proposed algorithm is implemented across 500 training episodes, and the average cost per iteration has been recorded. 
The training processes are recorded in Figure \ref{figure2new1}, which shows the average cost occurs per iteration, where we observe a stable cost at the end of the training horizon. Thus, the trained optimal policies are also stable. 
Next, we compare the proposed algorithm with $Q$-learning and classic Dyna-$Q$ as benchmarks. All three algorithms go through the same training phase with horizon $T = 100$ days. Then, the trained $Q$-value and estimated optimal policy are applied for testing in the same environment, with the average daily cost within another 100 days.

\begin{table*}[htbp]
\centering
\caption{Costs on testing set and training time consumed} 
\label{tab:costs_and_training_time}
\begin{tabular}{@{}lcccccc@{}} 
\toprule
\text{Dataset} & {\makecell[c]{ Model}}& \text{Algorithm} & {\makecell[c]{ Average\\ daily cost}} & {\makecell[c]{ Cost\\ improvement}} &{\makecell[c]{Training time (s)\\ per episode}} &{\makecell[c]{Training time\\ improvement}} \\

\midrule
\multirow{6}{*}{$\sigma^2 = 1$} & \multirow{3}{*}{MLP} & Adjusted Dyna-$Q$ & 1.614 & --- & 494.33 & 76.35\% \\
                       &                      & Dyna-$Q$  & 1.099 & 12.92\% & 2090.42 & --- \\
                       &                      & $Q$-learning & 1.262 & --- & 0.50 & --- \\
\cline{2-7}

                       & \multirow{3}{*}{BNN} & Adjusted Dyna-$Q$ & 1.054 & 23.40\% & 1043.35 & 67.87\% \\
                       &                      & Dyna-$Q$  & 0.922 & 33.00\% & 3247.64 & --- \\
                       &                      & $Q$-learning & 1.376 & --- & 0.50 & --- \\
\midrule
\multirow{6}{*}{$\sigma^2 = 3$} & \multirow{3}{*}{MLP} & Adjusted Dyna-$Q$ & 1.892 & 9.65\% & 490.85 & 76.73\% \\
                       &                      & Dyna-$Q$  & 1.908 & 8.88\% & 2109.79 & --- \\
                       &                      & $Q$-learning & 2.094 & --- & 0.49 & --- \\
\cline{2-7}
                       & \multirow{3}{*}{BNN} & Adjusted Dyna-$Q$& 1.785 & 23.65\% & 777.15 & 77.54\% \\
                       &                      & Dyna-$Q$  & 1.822 & 22.07\% & 3460.02 & --- \\
                       &                      & $Q$-learning & 2.338 & --- & 0.50 & --- \\
\midrule
\multirow{6}{*}{$\sigma^2 = 5$} & \multirow{3}{*}{MLP} & Adjusted Dyna-$Q$& 2.035 & 15.24\% & 492.63 & 76.18\% \\
                       &                      & Dyna-$Q$  & 2.286 & 4.79\% & 2068.46 & --- \\
                       &                      & $Q$-learning & 2.401 & --- & 0.50 & --- \\
\cline{2-7}
                       & \multirow{3}{*}{BNN} & Adjusted Dyna-$Q$& 1.914 & 19.95\% & 862.54 & 73.23\% \\
                       &                      & Dyna-$Q$  & 2.022 & 15.43\% & 3222.19 & --- \\
                       &                      & $Q$-learning & 2.391 & --- & 0.50 & --- \\
\bottomrule
\end{tabular}

\end{table*}

The performance comparison of adjusted Dyna-$Q$, classic Dyna-$Q$, and $Q$-learning algorithms is shown in Table \ref{tab:costs_and_training_time}. The average cost during the testing period by $Q$-learning is used as the comparison basis for investigating the improvements by the other two approaches. It can be observed that the model by BNN can indeed produce a lower average daily cost than the model by MLP, but with a relatively higher training time, across all three algorithms when $\sigma^2 =3,\,5$. The results indicate that the adjusted Dyna-$Q$ algorithm can reduce the training time per episode of classic Dyna-$Q$ by 77.5\% while offering a cost reduction of up to 23.7\% for testing. 

Next, we implement the adjusted Dyna-$Q$ algorithm with transfer learning. We investigate the following two scenarios, training and testing, respectively. Here, the bakery manager is required to form an ordering policy at an early stage after the new product is released, without prior knowledge of its demand distribution. We choose the parameters $\alpha=0.1,\,\gamma=0.9,\,\partial_{\epsilon}=1000,\,\partial_{N} = 1000$ and $\sigma^2 = 5$.

\textbf{Scenario 1 (Training):} In this scenario, we set: $\epsilon_0 = 0.4,\,\epsilon_{\min} = 0,\,N_0 = 100,\,N_{\min} = 0$. We first implement the method in Section II.$B$ with $h= 10$ and incorporate transfer learning into the model of adjusted Dyna-$Q$. Then the training starts right at iteration $t=1$, when the new product, Boule 400g, has just been released and its demand is observed. The training horizons are set as one month i.e., $T=30$. The total cost of the training will be recorded, and its average value will be over 100 training episodes. We also record the variance of total cost over all episodes.
Table \ref{Table 4.7 Results on the realistic environment (first month)} records the experiment results where the ``Shortage
percentage'' records the proportion of days within a month when the unfulfilled demand occurs.  
\begin{table}[h!] 
\centering 
\caption{Results for training (\textbf{Scenario 1)}} 
\label{Table 4.7 Results on the realistic environment (first month)}
\begin{tabular}{@{}lcccccc@{}} 
\toprule 
\text{Algorithm} & \text{\makecell[c]{Transfer\\learning}} & {\makecell[c]{ Average \\ total \\cost}} & {\makecell[c]{ Shortage \\percentage}} &{\makecell[c]{ Daily\\average\\inventory \\holding}} &{\makecell[c]{ Variance \\in total\\ cost}} \\ 
\midrule
\multirow{2}{*}{\makecell[c]{Adjusted \\Dyna-$Q$} }
& $\surd$ & 62.36 & 0.308 & 5.0 & 40.4 \\ 
& $\times$ & 65.32 & 0.442 & 3.6 & 76.1 \\ 

\midrule 
\multirow{2}{*}{Dyna-$Q$} 
& $\surd$ & 66.63 & 0.271 & 5.9 & 28.0 \\ 
& $\times$ & 61.86 & 0.333 & 4.6 & 49.1 \\ 
\midrule 
\multirow{1}{*}{$Q$-learning} 
& $\times$ & 65.13 & 0.490 & 2.9 & 36.3 \\ 
\bottomrule 
\end{tabular} 
\end{table}

\textbf{Scenario 2 (Testing):}
In this scenario, we set: $\epsilon_0 = 0.3,\,\epsilon_{\min} = 0.1,\,N_0 = 20,\,N_{\min} = 10$, and run the training process the same as Scenario 1. After the training is completed, the trained $Q$-value and estimated optimal policy will be tested in the real environment for another month, and the total cost will be recorded. The testing will be repeated 100 times, and the average total cost and variance will be recorded. Table \ref{Table 4.9 Results on testing environment (second month)} records
the experiment results.

\begin{table}[h!] 
\centering 
\caption{Results for testing (\textbf{Scenario 2)}} 
\label{Table 4.9 Results on testing environment (second month)}
\begin{tabular}{@{}lcccccc@{}} 
\toprule 
\text{Algorithm} & \text{\makecell[c]{Transfer\\learning}} & {\makecell[c]{ Average \\ total \\cost}} & {\makecell[c]{ Shortage \\percentage}} &{\makecell[c]{ Daily\\average\\inventory \\holding}} &{\makecell[c]{ Variance \\in total\\ cost}} \\
\midrule
\multirow{2}{*}{\makecell[c]{Adjusted \\Dyna-$Q$} }
& $\surd$ & 65.74 & 0.227 & 6.5 & 93.8 \\ 
& $\times$ & 69.91 & 0.261 & 5.8 & 250.2 \\ 

\midrule 
\multirow{2}{*}{Dyna-$Q$} 
& $\surd$ & 83.52 & 0.385 & 5.1 & 30.8 \\ 
& $\times$ & 74.32 & 0.317 & 5.7 & 133.3 \\ 
\midrule 
\multirow{1}{*}{$Q$-learning} 
& $\times$ & 74.29 & 0.150 & 8.4 & 21.2 \\ 
\bottomrule 
\end{tabular} 
\end{table}

From the experimental results of Scenario 1 recorded in Table \ref{Table 4.7 Results on the realistic environment (first month)}, regarding the total inventory cost generated in the first month, the three algorithms do not show a distinct difference in Scenario 1, with or without transfer learning. Note that the basic Dyna-$Q$ is the most superior, with the total cost during the training period at 61.86. However, incorporating transfer learning helps to stabilize all the algorithms through total cost variance reduction by at least $31.6\%$. From the experimental results of Scenario 2 recorded in Table \ref{Table 4.9 Results on testing environment (second month)}, we observe that the adjusted Dyna-$Q$ algorithm with or without transfer learning performs better than all the other algorithms. In addition, incorporating transfer learning helps the adjusted Dyna-$Q$ algorithm perform the best on testing. Another observation is that the algorithms with transfer learning perform more stably with significantly lower variance in total cost on testing, than those without transfer learning. Finally, we observe the phenomenon that the adjusted Dyna-$Q$ algorithm produces policies reducing the shortage percentage by holding more inventory on average. 

\begin{figure}[htbp]
\centering
\includegraphics[scale=0.37]{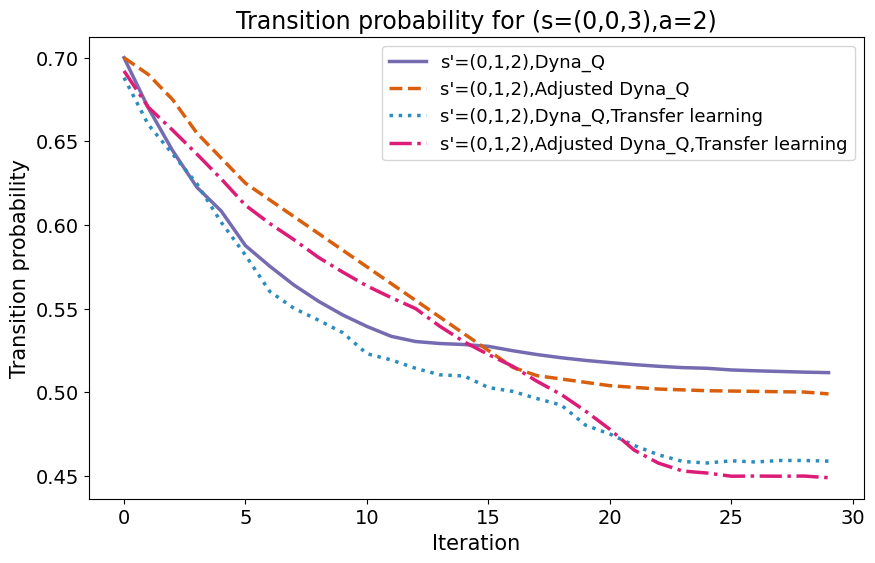}
\caption{Transition probability comparison} 
\label{figure3_transition}
\end{figure}

Compared with the results in Scenario 1, the variance of the total cost is much more significantly reduced in Scenario 2, which shows the strong out-of-sample performance of the proposed approach and its strong reliability in real-life applications. From both Table \ref{Table 4.7 Results on the realistic environment (first month)} and \ref{Table 4.9 Results on testing environment (second month)}, we also observe that the inclusion of
transfer learning actually increases the average training
cost for the classic Dyna-Q algorithm. This suggests that transfer learning may
introduce early-stage bias, and such bias will be aggregated during long and fixed planning steps, when the agent lacks mechanisms
(like STC) to correct for model mismatch. While, the adjusted Dyna-$Q$ indicates a significant effect in total cost reduction due to the decaying exploration probability and planning steps that can eliminate the discrepancy between the model and the new product's demand distribution. \WJmodified{In Figure \ref{figure3_transition}, we monitor the transition probability of one state-action pair in the $M(s,\,a)$ with $s = (0,\,0,\,3)$, $a=2$ and $s^\prime = (0,\,1,\,2)$, in 30 iterations. In this case, the observed daily demand is implied as $2$ and its true probability is $0.12$. It can be found that the transition probability obtained by the Adjusted Dyna-$Q$ algorithm with transfer learning is the closest to the true probability. The improvements brought both by adjusted Dyna-$Q$ and transfer learning are also shown. }

\section{Conclusion}
The paper presents a new RL algorithm for inventory management of the newly launched products where historical demand information is unavailable. The proposed adjusted Dyna-$Q$ algorithm is built upon the classic Dyna-$Q$ algorithm, aiming to strike a balance between model-based and model-free approaches during training. Additionally, based on the idea of transfer learning, a demand forecasting model is introduced, leveraging historical demand data from existing similar products to provide warm-start information for the adjusted Dyna-$Q$ algorithm.

A case study using real-life data is conducted to validate the proposed approach. The adjusted Dyna-$Q$ algorithm demonstrates significant benefits, including up to a 23.7\% reduction in average daily cost compared to $Q$-learning and up to a 77.5\% decrease in training time compared to classic Dyna-$Q$. Training and testing performances of incorporating generating transfer learning or not into the algorithm are shown and compared. The results show that the adjusted Dyna-$Q$ algorithm achieves the lowest total cost and relatively low shortage percentages compared to other algorithms. Algorithms that incorporate transfer learning exhibit greater stability and lower variance in total cost during testing.

For future research directions, several possibilities are suggested. First, further applications of the proposed approach in various inventory management contexts and industries could assess its versatility and scalability. Second, evaluating the adaptability of the approach to specific uncertainty patterns, such as seasonal demand fluctuations or sudden market changes, could provide insights into its robustness. Lastly, enhancing the demand prediction model to measure similarity between existing products and the new product could improve the incorporation of historical data from similar products with high similarity during training.







\bibliographystyle{IEEEtran}
\bibliography{IEEEabrv,reference}

\end{document}